\documentclass[letterpaper, 10 pt, conference]{ieeeconf}  
\IEEEoverridecommandlockouts                              
\pdfoutput=1
\usepackage{color}
\usepackage{graphicx}
\usepackage{crop}
\usepackage{multirow} 
\usepackage{caption}
\usepackage{amsmath}
\usepackage{amssymb}
\usepackage{xspace}
\usepackage{subfig}
\usepackage{booktabs}
\usepackage{dsfont}

\usepackage[export]{adjustbox}
\newcommand{\set}[1]{ {\mathcal #1}\xspace}

\def\eg{\textit{e.g.}\xspace}
\def\ie{\textit{i.e.}\xspace}

\def\aka{\textit{a.k.a.}\xspace}

\title{\LARGE \bf
Robust Place Categorization with
Deep Domain Generalization
}

\author{Massimiliano Mancini$^{1,2}$, Samuel Rota Bul\`o$^{2,3}$, Barbara Caputo$^{1}$, Elisa Ricci$^{2,4}$
\thanks{This work was partially supported by the ERC grant 637076 - RoboExNovo (B.C. ), and the CHIST-ERA project ALOOF (B.C.).}
\thanks{$^{1}$M. Mancini and B. Caputo are with University of Rome La Sapienza, Rome, Italy.
        {\tt\small \{mancini,caputo\}@dis.uniroma1.it}}%
\thanks{$^{2}$M. Mancini, S. Rota Bul\`o and E. Ricci are with Fondazione Bruno Kessler, Trento, Italy. {\tt\small \{rotabulo,eliricci\}@fbk.eu}}%
\thanks{$^{3}$S. Rota Bul\`o' is with Mapillary Research, Graz, Austria.}
\thanks{$^{4}$E. Ricci is with University of Perugia, Perugia, Italy.}%
}

\begin{document}

\begin{titlepage}
\null
\vfill
\renewcommand{\fboxsep}{10pt}
\fbox{\Large\begin{minipage}{\columnwidth}
\textbf{Disclaimer:}

This work has been accepted for publication in the IEEE Robotics and Automation Letters:\vspace{4pt}
\newline
doi:     10.1109/LRA.2018.2809700
\newline
link:    https://ieeexplore.ieee.org/document/8302933/
\newline
\newline
\textbf{Copyright:} 
\newline
\copyright~2018 IEEE. Personal use of this material is permitted. Permission from IEEE must be obtained for all other uses,  in  any  current  or  future  media,  including  reprinting/  republishing  this  material  for  advertising  or promotional purposes, creating new collective works, for resale or redistribution to servers or lists, or reuse of any copyrighted component of this work in other works.
\newline
\end{minipage}}
\vfill
\clearpage
\end{titlepage}
\maketitle
\begin{abstract}

Traditional place categorization approaches in robot vision assume that training and test images have similar visual appearance. Therefore, any seasonal, illumination and environmental changes typically lead to severe degradation in performance. To cope with this problem, recent works have proposed to adopt domain adaptation techniques. While effective, these methods assume that some prior information about the scenario where the robot will operate is available at training time. Unfortunately, in many cases this assumption does not hold, as we often do not know where a robot will be deployed. To overcome this issue, in this paper we present an approach which aims at learning classification models able to generalize to unseen scenarios.
Specifically, we propose a novel deep learning framework for domain generalization. 
Our method develops from the intuition that, given a set of different classification models associated to known domains (\eg corresponding to multiple environments, robots), the best model for a new
sample in the novel domain can be computed directly at test time by optimally combining the known models. 
To implement our idea, we exploit recent advances in deep domain adaptation and design a Convolutional Neural Network architecture with novel layers performing a weighted version of Batch Normalization. 
Our experiments, conducted on three common datasets for robot place categorization, confirm the validity of our contribution.

\end{abstract}

\section{INTRODUCTION}
\label{intro}

In recent years we have witnessed great advancements in computer and robot vision thanks to deep learning models. 
In particular, convolutional neural networks (CNN) have reached outstanding performances in many different tasks such as object classification \cite{he2016deep}, depth estimation \cite{xu2017multi} and affordance prediction \cite{porzi2017learning}. 
Despite their effectiveness, CNNs have some drawbacks. First, they are data-hungry, \ie very large labeled dataset are usually required for training. This is a major issue in robotics, where data acquisition and annotation is especially time consuming and often infeasible. Second, most CNN 
models 
are derived assuming that training and test data belong to the same distribution. This is a clear limitation for robots operating in the real world. 
In fact, robots should be able to function in completely different places, under many diverse environmental conditions: therefore, they cannot simply employ pre-trained models unable to generalize to arbitrary settings. 

This work focuses on the problem of semantic place categorization from visual data, where these issues are especially relevant \cite{wu2009visual}. Correctly 
identifying the semantic category of a place is important for robot localization, mapping and exploration \cite{stachniss2006speeding,kostavelis2015semantic}. Unfortunately, environmental changes (\eg due to the presence of people or obstacles and to changing lighting conditions) 
make this task extremely challenging.  

Traditional place categorization approaches \cite{wu2011centrist,fazl2012histogram,urvsivc2016part,mancini2017learning} require  labeled datasets of training images. 
While the resulting models are very accurate when test samples are similar to training data, 
their performance significantly degrade when the robot collects images with very different visual appearance \cite{pronobis2010realistic}. To address this issue, recent works have exploited \emph{domain adaptation (DA)} techniques \cite{prasath2012transfer,costante2013transfer,kira2014transfer}. These methods develop models which are meant to be effective in the scenario where the robot will operate, \ie the \emph{target} domain. However, since few target data are often available, DA attempts to transfer useful knowledge
from a larger set of \textit{source} data (\eg previously collected by other robots or derived from publicly-available datasets).
 

\begin{figure}[t]
\centering
\includegraphics[width=0.95\columnwidth,trim={0 0.75cm 3.5cm 0},clip]{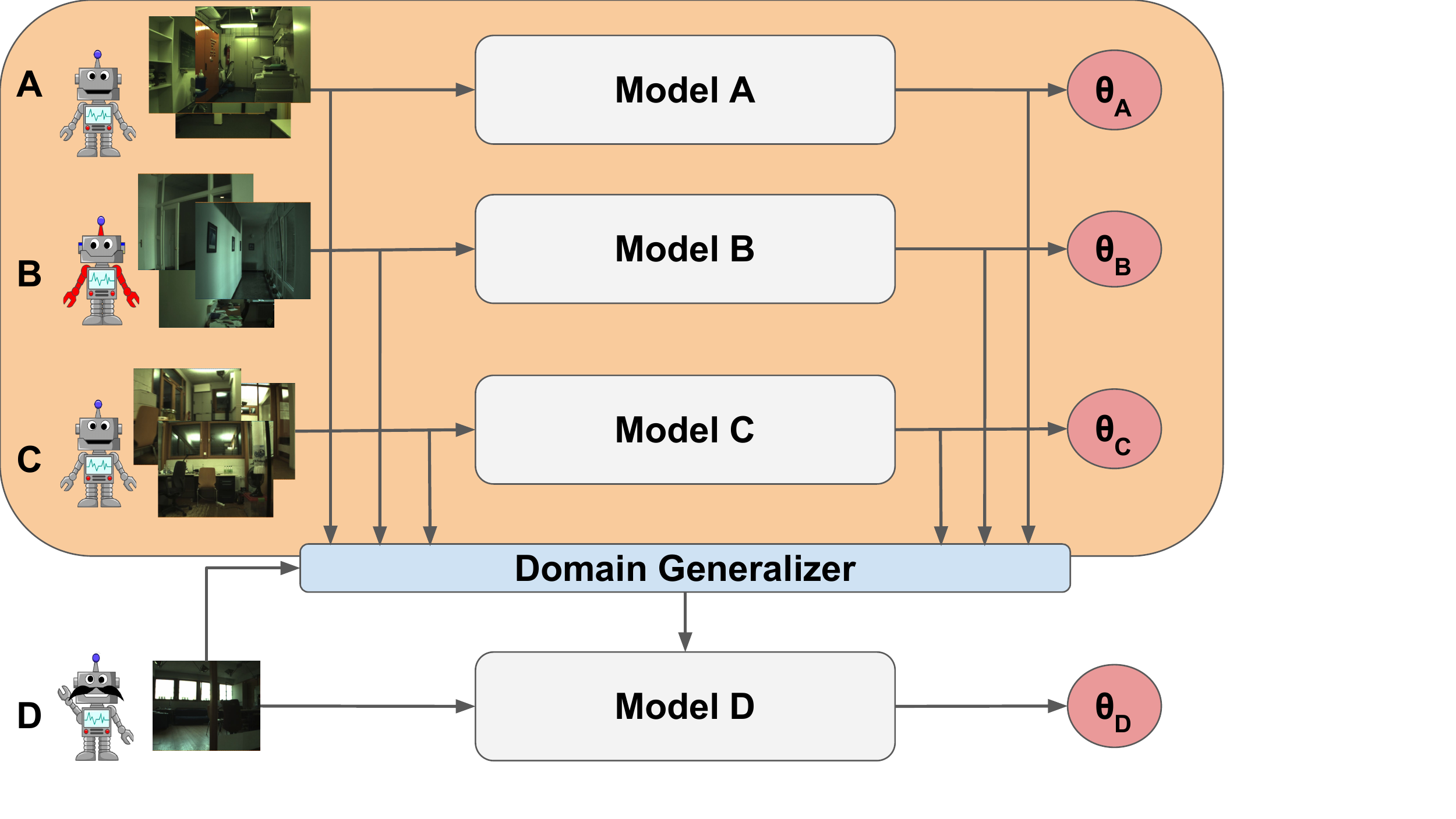}
\caption{The domain generalization problem. At training time (orange block) images of multiple source domains (\eg 
A,B,C) are available. These images are used to train different models with parameters $\theta_i$. 
Our approach automatically computes a model D which accurately classifies images of a novel domain (not available during training)
by combining the models of the known domains.}
   \label{fig:teaser}
   \vspace{-20pt}
\end{figure}


While domain adaptation algorithms provide effective solutions, they require some prior knowledge of the target domain  at training time, \eg to have access to target data. 
Unfortunately, this information may not always be available. Consider for instance an household robot: since the number of possible customers is huge, it is inconceivable to collect data for each possible house and application scenario. 
In this paper, we argue that a more relevant problem in the context of semantic place categorization is \emph{domain generalization (DG)}. Opposite to DA, where target data are exploited to produce a classifier accurate under specific working conditions, the idea behind DG is to learn a domain agnostic model applicable to any unseen target domain. In other words, in this paper we  are interested in learning a place categorization model which is 
as general as possible and employable by different robots and in various environmental conditions. 

While many strategies are possible for addressing domain generalization \cite{muandet2013domain,khosla2012undoing}, in this paper we propose a novel deep learning framework. Our approach develops from the idea that, given data from multiple source domains and the associated models, 
the best model for the target domain can be generated on-the-fly when a novel sample arrives by optimally combining the precomputed models from source domains (see Fig.\ref{fig:teaser}). To implement this idea we design a novel CNN architecture which relies on two main components. First, inspired by recent works on domain adaptation \cite{carlucci2017just,li1603revisiting}, we construct multiple source models by embedding into a common CNN few domain-specific Batch Normalization layers. In this way, different classifiers can be built keeping the number of parameters limited. 
Second, we design a lateral network branch which computes the likelihood that a certain instance belongs to a given domain. When applied to a novel target sample, this branch calculates 
its 
probabilities to be part of the different source domains. These values are used to construct the target classifier performing a combination of known source models. To this aim, the novel Weighted Batch Normalization (WBN) layers are introduced. We demonstrate the effectiveness of the proposed DG approach with extensive experiments on three 
datasets, namely the COsy Localization Database (COLD) \cite{pronobis2009ijrr}, the Visual Place Categorization (VPC) dataset \cite{wu2009visual} {and the Specific PlacEs Dataset (SPED) \cite{chen2017deep}}. Moreover, we show how the proposed framework can be employed where no prior information about source domains is available at training time: given a training set, our model can be used to automatically cluster training data and learning multiple models, discovering latent domains and associated classifiers.

To summarize, the main contribution of this paper is twofold. 
Firstly, we introduce the problem of domain generalization for semantic place recognition and we present the first deep learning framework for addressing it. Secondly, we show how explicitly embedding domain information into learning is 
beneficial and permits to significantly increase classification accuracy over 
domain-agnostic approaches.

\section{RELATED WORK}
\label{related_work}

Devising models which are robust to domain shift is of fundamental importance for developing robotic systems working in the wild. 
Therefore, several recent works have focused on learning
models which are as much as possible invariant to seasonal \cite{kanji2015cross,viswanathan2016vision}, illumination \cite{carlevaris2014learning,lu2015robustness,mount20162d,upcroft2014lighting} and environmental  \cite{milford2012seqslam,tsukamoto2015self} changes. While several tasks such as place recognition \cite{kanji2015cross}, SLAM \cite{milford2012seqslam}, semantic segmentation \cite{upcroft2014lighting} benefit from these type of models, 
in this paper we specifically consider the problem of semantic place categorization from visual data. 


Standard approaches addressing this task extract global features,  either hand-crafted \cite{wu2011centrist, pronobis2006discriminative, fazl2012histogram} or CNN-based \cite{zhou2014learning}, from the input images and subsequently train a classifier to discriminate among a given set of categories. Other methods consider extracting local features, either through a fully-convolutional neural network \cite{mancini2017learning} or regions of interest pooling \cite{urvsivc2016part}. Differently, 
some approaches attempt to classify places analyzing the occurrences of specific objects and derive 
object-level information either adopting CNNs \cite{liao2016understand} or object templates \cite{yang2012object}. However, none of these methods is effective when tested in cross-domain conditions (see for instance experimental results in \cite{mancini2017learning, fazl2012histogram,yang2012object}). 

To address the problem of robustness to varying environmental conditions, 
some works 
considered 
domain adaptation and transfer learning techniques.
For instance, \cite{luo2007svm} proposes an approach based on adaptive support vector machine. 
In \cite{prasath2012transfer} a supervised transfer learning approach based on Least Square SVM is described. In \cite{costante2013transfer} an unsupervised technique selects the source data useful for place recognition in the target domain. Similarly, \cite{kira2014transfer} introduces a transfer learning approach based on sparse coding. 
Despite their effectiveness, all these methods assume the availability of some information for the target domain. This is a clear limitation, as in many cases a robot operates in completely unknown scenarios. Therefore, our work specifically tackles the problem of domain generalization.  
To our knowledge, no previous works have proposed a DG method for robot place categorization.

Domain generalization \cite{muandet2013domain} has been recently studied in the literature, \eg in the context of visual object recognition
and action classification \cite{xu2014exploiting,khosla2012undoing}. However, most previous works adopt hand-crafted features and, to our knowledge, this is the first method proposing an end-to-end deep learning architecture for DG in robot-vision.
\vspace{-2pt}

\section{Domain Generalization with Weighted Batch Normalization}

In this section we present our novel approach for domain generalization. 
We first focus on the closely-related problem of domain adaptation, briefly describing a recent deep learning model \cite{carlucci2017just} for coping with domain shift by adopting a revised version of Batch Normalization (BN) \cite{ioffe2015batch}. We then show how a similar strategy can be employed to address the more challenging problem of domain generalization, introducing the proposed Weighted Batch Normalization layers.


\subsection{Domain Adaptation through Batch Normalization}
Several previous works have considered the problem of domain adaptation in the context of semantic place categorization \cite{luo2007svm,costante2013transfer}. Indeed, modern robotic platforms require classification models which are robust to variations in the environmental conditions. Unfortunately, even state-of-the-art methods based on powerful CNNs \cite{mancini2017learning,urvsivc2016part} tend to loose effectiveness when changes in illumination conditions, sensors or environments occur. Therefore, recent works have focused on developing domain adaptation methods for deep architectures \cite{DBLP:journals/corr/WulfmeierBP17}. The unsupervised DA problem is especially relevant in robotics, as annotating data is often practically infeasible when a robotic platform is deployed in a novel scenario.
Formally, this problem can be stated as follows. Given
a set of labeled images $\mathcal{X}_s=\{(I_1^s,y_1), \dots , (I_{n_s}^s,y_{n_s})\}$ from a source domain (\aka the \textit{source} set), where $I^s_i$ denote the images collected by a robot or obtained from publicly-available datasets and $y_i \in \{1, \dots, K\}$ are the labels indicating the rooms types (\eg corridor, office, \dots), and an unlabeled set of images $\mathcal{X}_t=\{I_1^t, \dots, I_{n_t}^t\}$  from a target domain (\aka \emph{target} set) corresponding to visual data collected by a robot in the novel scenario, we are interested in learning a classification model from $\mathcal{X}_s$ and $\mathcal{X}_t$ which accurately classify the target data. 



Recent works \cite{li1603revisiting,carlucci2017just,carlucci2017autodial} have shown that an effective strategy for unsupervised DA consists in embedding into CNN architectures specific Domain Alignment Batch Normalization (DA-BN) layers.
DA-BN layers are derived from the common BN technique \cite{ioffe2015batch} adopted for avoiding internal covariate shift within deep neural networks. 

\begin{figure}[t]
  \centering
  \subfloat[AlexNet+BN]{\includegraphics[width=0.9\columnwidth,trim={0 5cm 0 4cm},clip]{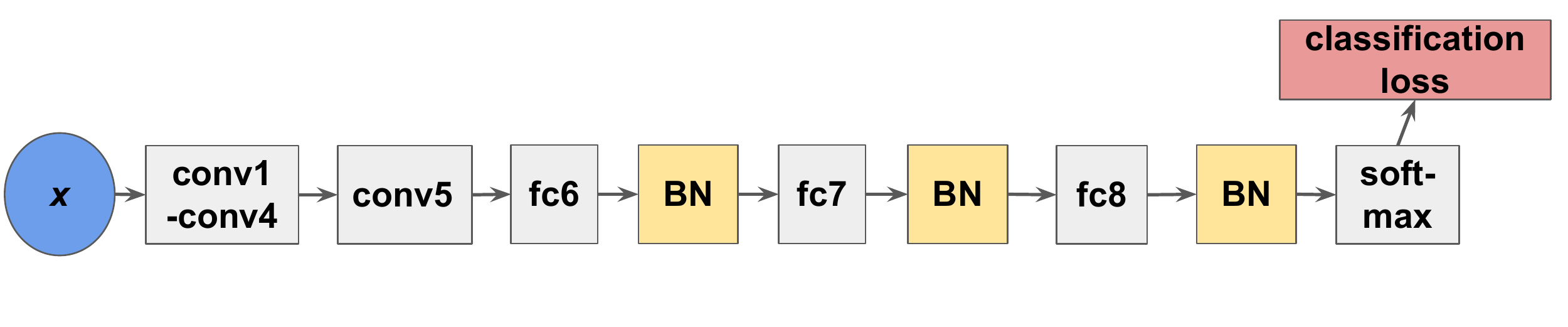}\label{fig:bn}}
  \hfill
  \subfloat[AlexNet+DA-BN]{\includegraphics[width=0.9\columnwidth,trim={0 4cm 0 3.2cm},clip]{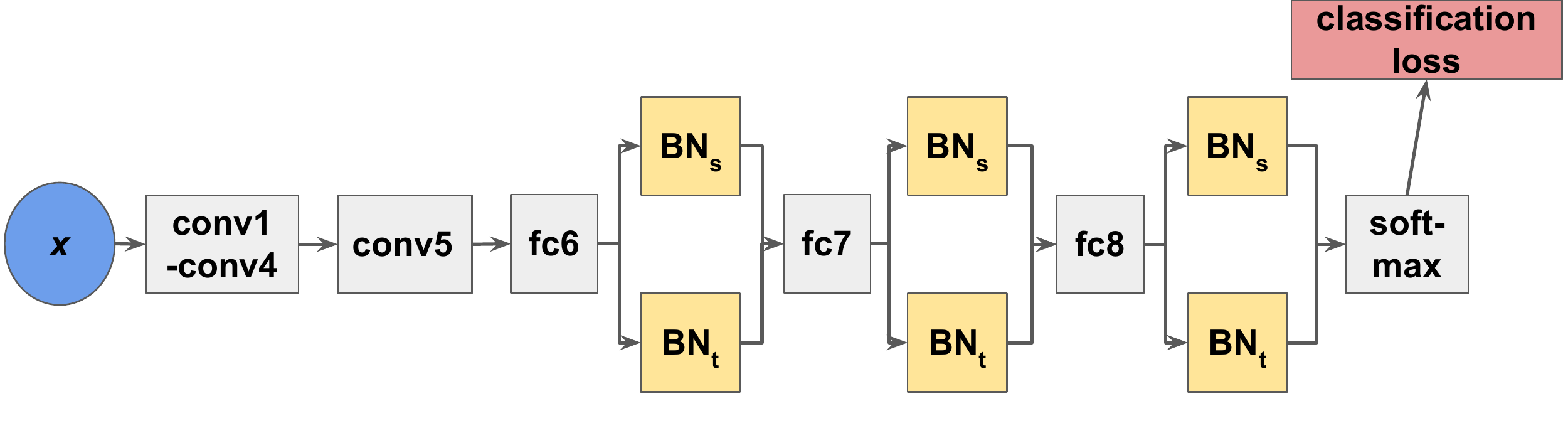}\label{fig:da-bn}}
  \hfill
  \subfloat[AlexNet+WBN]{\includegraphics[width=0.9\columnwidth,trim={0 1.2cm 0 0},clip]{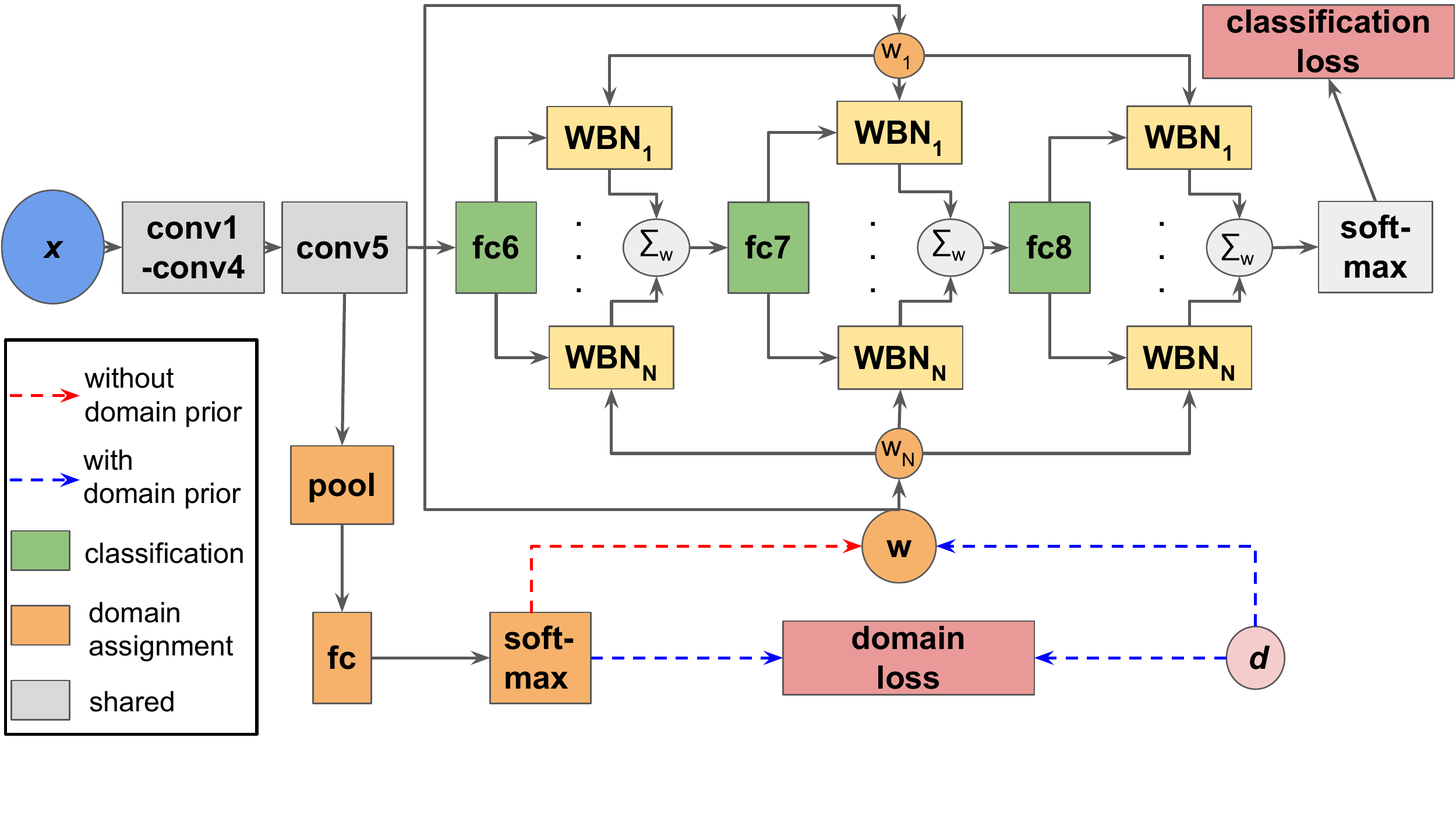}\label{fig:wbn}}
  \caption{Example of the proposed framework. (a) AlexNet with BN layers after each fully connected. (b) The same network employing DA-BN for domain adaptation, where different BN are used for source ($BN_s$) and target ($BN_t$). (c) Our approach for DG with WBN layers. 
  }
  \label{fig:method}
  \vspace{-20pt}
\end{figure}

The principle of BN is simple: the input distribution of a layer is kept constant by imposing a normalization of the input features. 
Denoting by $x_{i}$ the input of BN for a given feature channel and spatial location, 
BN operates by replacing $x_{i}$ with:
\begin{equation}
\label{eq:bn}
 \hat{x}_{i}=\gamma\frac{x_{i}-\mu_{\mathcal{X}}}{\sqrt{\sigma_{\mathcal{X}}^2 + \epsilon}} + \beta\,,
\end{equation}
where $\gamma$ is a scale factor and $\beta$ is a bias term. The components $\mu_{\mathcal{X}}$ and $\sigma_{\mathcal{X}}^2$ are respectively the mean and the variance computed over the training set $\mathcal{X}$, w.r.t. the chosen layer and feature. At training time, the statistics are approximated through the samples of the current batch, while at test time the global estimation is used. Figure \ref{fig:bn} illustrates a common CNN with BN layers inserted after each fully-connected.

The main idea behind the design of DA-BN layers is to perform the normalization using domain-dependent statistics. 
In this way, the alignment between different domains is realized by forcing source and target feature distributions to match the same reference distribution, a standard Gaussian 
in this case. Formally, samples from the source and the target domains are normalized according to their associated statistics $\{\mu_{\mathcal{X}_j},\sigma_{\mathcal{X}_j}^2\}$, $j \in \{s, t\}$, \ie:
\begin{equation}
\label{eq:da-bn-s}
 \hat{x}_{i}^j=\gamma\frac{x_{i}^j-\mu_{\mathcal{X}_j}}{\sqrt{\sigma_{\mathcal{X}_j}^2 + \epsilon}} + \beta\,.
\end{equation}
A key aspects of the methods in \cite{carlucci2017just,carlucci2017autodial} is to implement two different predictors for source and target data without requiring additional domain-specific parameters. In other words, 
DA-layers are embedded in a standard architecture, \eg AlexNet, and the same network parameters $\theta$ are shared by the source and the target predictors. During training, $\theta$ is computed minimizing a classification loss (see Fig. \ref{fig:da-bn}). It is worth noting that, analogously to BN, DA-BN layers can be embedded into many different types of architectures.
%


\subsection{Domain Generalization with Weighted BN}
As stated in Section \ref{intro}, one issue with DA methods is that they require the presence of a target set $\mathcal{X}_t$ in the training phase. This implies that data collected by a robot in the scenario of interest should be available for learning the classification model. However, a more realistic situation is when a robot is employed in a completely unseen environment. As an example, consider a service robot: it is unfeasible to collect data for all possible working environments. Therefore, in this work we focus on domain generalization.

\subsubsection{Weighted Batch Normalization for DG}
\label{sec:dg-hard}Formally, the DG problem can be stated as follows.
Suppose we have a set $\mathcal{X}=\{\mathcal{X}_1,\dots,\mathcal{X}_N\}$ corresponding to the union of data from $N$ source domains. The source sets correspond, \eg, to data acquired by multiple robots in different environments. We can write $\mathcal{X}=\{(I_1,y_1, d_1), \dots, (I_{n},y_{n},d_n)\}$, where $I_i$ is an image, $y_i$ its corresponding label and $d_i \in \{1, \dots, N\}$ is the domain 
to which the image belongs. No data are available for the target domain. DG aims at learning a model 
from $\mathcal{X}$ which performs well on previously unseen target data.

Given the data from the source domains a set of domain-specific classifiers can be obtained by 
extending \eqref{eq:da-bn-s} to a multi-domain formulation:
\begin{equation}
\label{eq:dg-hard}
\hat{x}_i =\gamma\sum_{j=1}^N\mathds{1}_{d_i=j}\frac{x_i-\mu_{\mathcal{X}_j}}{\sqrt{\sigma_{\mathcal{X}_j}^2 + \epsilon}} + \beta\,,
\end{equation}
where 
the statistics $\{\mu_{\set X_j},\sigma_{\set X_j}\}$ are specific for the domain $\mathcal{X}_j$ and $\mathds{1}_{d_i=j}$ is an indicator function that gives value 1 if $d_i=j$ and 0 otherwise. 

While knowledge about domains is available at training time for source data, we do not know a priori the domain of a target sample at test phase. To solve this problem, we propose to replace the hard assignment of \eqref{eq:dg-hard} with a soft assignment. Given a set of weights $w_{i,j}$, we perform feature normalization with Weighted Batch Normalization layers:
\begin{equation}
\label{eq:dg-soft}
\hat{x}_i =\gamma\sum_{j=1}^{N}w_{i,j}\frac{x_i-\mu_{\mathcal{X}_j}}{\sqrt{\sigma_{\mathcal{X}_j}^2 + \epsilon}} + \beta\,,
\end{equation}
where $\sum_{j=1}^{N}w_{i,j} = 1$ and $\forall j\; w_{i,j}\geq 0$. 
{The intuition behind this choice is deriving a classification model for the target domain as a combination of models from the source domains.}

In order to compute the weights $w_{i,j}$
, we propose to employ a separate network branch which originates from the first few convolutional layers of the main network (see Fig. \ref{fig:wbn}). This choice is motivated by the fact that end-to-end training is allowed and the number of parameters is kept limited. The specific architecture of the branch may be variable (see Sect. \ref{sec:implementation}), with the only restriction that its final output must be a probability vector of dimension $N$, corresponding to the number of known domains.

Analogously to what stated for DA-BN layers, the different source classifiers can be implemented by embedding WBN layers into a common CNN architecture. Let us denote by $\theta_s$ the parameters shared across the main network and the lateral branch. The network branch that produces the domain assignment weights has parameters denoted by $\theta_w$, while the parameters $\theta_c$ are shared across the domains but contribute only to the final classification and not to the assignment. 
During training we minimize the following loss:
\begin{equation}
\label{eq:stdloss}
L=\frac{1}{n}\sum_{i=1}^n \left[ \ell(x_i, y_i; \theta_c,\theta_s) + \lambda \ell_d(x_i, d_i; \theta_w,\theta_s) \right]\,.
\end{equation}
The loss is the sum of two terms, one considering place label information for accurate recognition, the other enforcing the lateral branch to successfully compute the correct domain.
\subsubsection{Discovering latent source domains} \label{sec:dg-soft}A drawback of the proposed DG approach is the need of multiple predefined source domains during training. 
In practice this information may not always be available and we may have access to a single training set.
However, even images from the same training set can have different visual appearance, \eg corresponding to multiple illumination conditions. In this case, it is reasonable to attempt to automatically discover the latent domains inside the training set and use this information in the proposed DG framework. 

Formally, in this scenario we have still a source dataset $\mathcal{X}$. 
However, the domain information is 
missing, \ie the source images $I_i$ do not have a domain label $d_i$ associated. 
Interestingly, our framework can be easily extended to this setting by entirely relying on the proposed soft assignment strategy. 
Our intuition is that, since similar input images will tend to produce similar outputs in the lateral network branch, implicitly visual data will be automatically clustered, enabling a latent domain discovery process. 

In this case, no domain loss $\ell_d$ is considered during training. 
Assuming the existence of $N$ latent domains, the statistics for the $j$-th domain 
can be computed by means of the weighting parameters $w_{i,j}$. In particular, at training time it is possible to approximately estimate the statistics by: 
\begin{align*}
\label{eq:wbn-stat}
\sigma_{\mathcal{X}_j}^2&=\sum_{i}^{n_b} 
\hat{w}_{i,j} (x_i-\mu_{\mathcal{X}_j})^2\\
\mu_{\mathcal{X}_j} &= \sum_{i}^{n_b} 
\mu_{\mathcal{X}_j} &= \sum_{i}^{n_b} 
\hat{w}_{i,j} x_i
\end{align*} 
with
$\hat{w}_{i,j}=\frac{w_{i,j}}{\sum_{k}^{n_b} 
w_{k,j}}$
and $n_b$ is the number of samples in the current batch. 
As for standard BN, these values 
update the global estimate of the statistics employed at test time.
Since the definition of domain can be subtle 
and the number of domains unknown, we believe that this formulation provides a more generic framework for addressing DG.


\subsubsection{Implementation details} 
\label{sec:implementation}
To implement the proposed DG framework, 
similarly to DA-BN, we replace each BN with a number of WBN layers equal to the domains available for training. For computing the assignment weights 
the lateral branch takes as input the last frozen layer, applying a ReLU non linearity followed by a fully-connected layer and a softmax activation. 
{In case the input is taken from a convolutional layer, a global pooling is applied before the lateral branch. In the large scale experiments on SPED, the lateral branch takes as input the first convolutional layer but two convolutional layers are added before the global pooling. These layers have the same parameters of \texttt{conv2} and \texttt{conv3} of the main network.}
From the assignment, we perform a weighted normalization of the input, summing the results obtained from all the domains. The scale $\gamma$ and the bias $\beta$ parameters are shared across the domains.
For the loss functions $\ell$ and $\ell_d$ we choose a log-loss, with the input normalized through the softmax operator.

\section{EXPERIMENTAL RESULTS} 
In this section we demonstrate the effectiveness of our domain generalization approach for semantic place categorization. We first perform a preliminary experiment to show that, when samples from the target domain are available (\textit{i.e.} we consider a domain adaptation setting), the adoption of domain-specific batch normalization layers into a CNN as in \cite{carlucci2017autodial,carlucci2017just} is beneficial and guarantees a significant improvement in terms of recognition accuracy (Subsection ~\ref{exp:domain}). In Subsection~\ref{exp:generalization} we test the proposed approach in the more challenging task of domain generalization and compare it with state-of-the-art methods, {while in Subsection~\ref{exp:large-scale} we provide further results in a large scale outdoor scenario.}\vspace{-5pt}

\subsection{Experimental settings}
\label{exp:setting}
\subsubsection{Datasets} In our experiments we use three robot vision datasets, namely the widely adopted COLD \cite{pronobis2009ijrr} and VPC  \cite{wu2009visual} datasets, {and the recent SPED dataset \cite{chen2017deep}}. 

The COLD Database contains three datasets of indoor scenes acquired in different laboratories and from different robots. The COLD-Freiburg (Fr) has 26 image sequences collected in the Autonomous Intelligent Systems Laboratory at the University of Freiburg, with a camera mounted on an ActivMedia Pioneer-3 robot. COLD-Ljubljana (Lj) contains 18 sequences acquired from 
an iRobot ATRV-Mini platform at the Visual Cognitive Systems Laboratory of University of Ljubljana. In the COLD-Saarbr\"ucken (Sa) an ActivMedia PeopleBot has been employed to gather 29 sequences inside the Language Technology Laboratory at the German Research Center for Artificial Intelligence in Saarbr\"ucken. 

The VPC dataset contains images acquired from several rooms of 6 different houses with multiple floors. The images are acquired by means of a camcorder placed on a rolling tripod, simulating a mobile robotic platform. The dataset contains 11 semantic categories, but only 5 are common to all houses: bedroom, bathroom, kitchen, living room and dining-room. Following previous works \cite{wu2009visual,fazl2012histogram,yang2012object}, we use the common categories in our experiments. 

{SPED is a large scale dataset introduced in the context of place recognition. It contains images of 2543 outdoor cameras collected from the Archive of Many Outdoor Scenes (AMOS) \cite{jacobs2007consistent} during February and August 2014\footnote{The full dataset is currently not available, but the authors provided us a subset with about 500 images per camera corresponding to 900 categories.}.

\subsubsection{Baseline architectures}{ For COLD and VPC} we perform experiments with two common architectures: AlexNet \cite{krizhevsky2012imagenet} and ResNet \cite{he2016deep}. For AlexNet we use the standard architecture pre-trained on Imagenet \cite{deng2009imagenet}. In all the experiments, we fine-tune the last two fully-connected layers, rescaling the input images to 227 $\times$ 227 pixels. For ResNet we consider the 10 layers version of the architecture, again pre-trained on ImageNet. In all the experiments, we rescale the input images to 224x224 pixels, fine-tuning the network starting from the last residual block.  Both the networks are trained with a weight decay of 0.0005 and an initial learning rate of 0.001, while the initial learning-rate of the final classifier is set to 0.01. The learning rate is dropped of a 0.1 factor after $90\%$ of the iterations. For the experiments on COLD, we use a batch-size of 256 for AlexNet and 64 for ResNet, 
training the networks for 1000 iterations. For VPC, we set the batch size to 128 and 64 for AlexNet and ResNet respectively, training the networks for 2000 iterations. The training parameters are the same for our method and the baselines and fine-tuning is performed for all the models.

The proposed approach can be applied to common CNNs by simply replacing standard BN layers with our WBN layers. While for ResNet BN layers are already employed, this is not true for AlexNet. For these experiments we employ a variant of AlexNet where BN layers are inserted after each fully-connected layer.

{For SPED we 
use AlexNet and the AMOSNet architecture, following \cite{chen2017deep}. AMOSNet is very similar to {AlexNet}, with the first fully-connected layer replaced by a convolutional layer and a pooling operation. We follow the same protocol of \cite{chen2017deep}, using the same hyper-parameters for training. 
We train both networks from scratch, applying BN or WBN layers after each layer with parameters, except the classifier. }  


The evaluation is performed using a NVIDIA GeForce 1070 GTX GPU, implementing all the models with the popular Caffe \cite{jia2014caffe} framework. For the baseline AlexNet architecture we take the pre-trained model available in Caffe, while for ResNet we consider the model from \cite{simon2016cnnmodels}. The code implementing the proposed method is publicly available\footnote{https://github.com/mancinimassimiliano/caffe}.

\subsection{Results}
\subsubsection{Domain Adaptation}
\label{exp:domain}

We first perform some preliminary experiments to demonstrate that applying domain-specific batch normalization layers as proposed in \cite{carlucci2017just} is beneficial for transferring domain knowledge in the context of robot place categorization.
Note that in this case, \textit{data from the target domain are available, either labeled or unlabeled}. As our goal is to demonstrate the
importance of source and target specific BN layers in our application scenario, we do not consider the entropy loss used in \cite{carlucci2017just} in our experiments. 

Following previous works on domain adaptation \cite{costante2013transfer}, we consider pair of sequences extracted from the COLD dataset. Similarly to \cite{costante2013transfer}, we take pairs where either different robots or 
lighting conditions are present\footnote{The pair containing sequence A2 of Saarbr\"ucken sunny has been excluded from our analysis as data are not available anymore.}. Specifically, we consider four pairs of videos: Freiburg Cloudy (Fr.C)-Saarbr\"ucken Cloudy (Sa.C), Freiburg Cloudy-Saarbr\"ucken Night (Sa.N), Ljubljana Sunny (Lj.S)-Freiburg Cloudy and Ljubljana Sunny-Saarbr\"ucken Night. As in \cite{costante2013transfer}, for Freiburg Cloudy and Ljubljana Sunny we use the standard sequence A1, for Saarbr\"ucken Cloudy the standard sequence A2 and for Saarbr\"ucken Night the extended sequence A1. 
With respect to \cite{costante2013transfer}, we focus on a slightly different scenario, where the two domains share common categories.  
In particular, we take the four common classes/rooms among all the different sequences: printer area, corridor, bathroom and office (obtained by merging 1-person and 2-persons office). 

We perform experiments in multiple settings, \ie considering labeled data from all the four categories and progressively discarding labels for the frames associated to certain categories. 
In practice, unlabeled images are still fed to the network (and contribute to the computation of BN statistics), but they do not have any loss term associated.
For each pair of sequences, we take 5 random splits, where $75\%$ of the data of each sequence are used for training and the other $25\%$ are used for testing. Inside the splits, all possible configurations are considered (\ie if for a sequence we test the configuration where 1 class lacks of labeled data, all the possible choices of that class are tested). 
Since the datasets are unbalanced, we report the results as average accuracy per class on the target domain, averaging the results across all the possible configurations and splits. 

Figure \ref{fig:graph-da} compares the results obtained using domain-specific BN layers (DA-BN), with those we get adopting the traditional AlexNet model (Base) and its batch-normalized version (BN). The figure clearly shows the advantages of exploiting domain priors when data for the target domain are available. This advantage is remarkable when the number of labeled classes for the target domain decreases. We ascribe this behavior to the fact that, by considering domain-specific statistics in BN, we are effectively normalizing the features of each domain, without recurring to a unified estimate and, implicitly, to an approximate normalization. Indeed, the normalization obtained with DA-BN produces an automatic feature alignment which helps the classifier to learn domain-independent prediction functions, increasing the generalization capabilities of the model. Notice that the increase in accuracy is less pronounced when only 1 labeled class is used in the target set. A possible reason is that in this case the unbalanced number of samples per class negatively affect the performance. 

 \begin{figure*}[t]
 \centering
 \hspace{-60pt}
\includegraphics[width=0.86\textwidth,trim={0cm 13cm 0cm 0cm},clip]{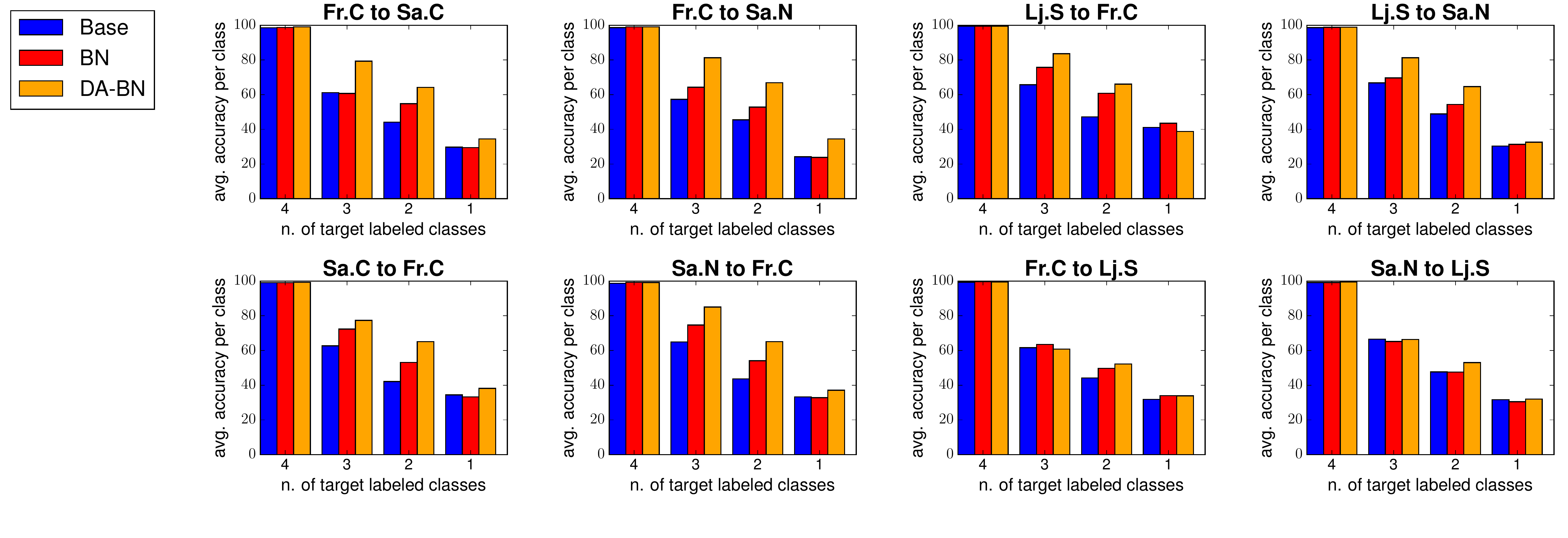}
\vspace{-15pt}
 \caption{Comparison between AlexNet (blue), AlexNet with BN (red) and DA-BN (yellow) on pair of sequences of COLD, varying the number of labeled classes in the target domain. A to B means we are using A as source domain and B as target.\vspace{-15pt}}
    \label{fig:graph-da}
 \end{figure*}

\subsubsection{Domain generalization}
\label{exp:generalization}
While in the previous experiments we exploit data from the target domain during learning, in this section we assume that the robot will be employed in an unseen environment, \ie we focus on the more challenging scenario of domain generalization. 
Our experiments aim to demonstrate the effectiveness of our approach in learning effective classification models in case of varying environmental conditions (\eg illuminations, laboratories).
In particular, we test two different variants of the proposed approach. In the first case (WBN$ ^*$) we consider the presence of domain priors at training time, as in Section \ref{sec:dg-hard}. In the second variant, WBN, we do not assume to have knowledge about domains at training time, thus our model just relies on the soft-assignment (Section \ref{sec:dg-soft}). 


We first perform experiments on the COLD database. For each laboratory and illumination condition we consider the standard sequences 1 of part A, except for Saarbr\"ucken Cloudy, for which we take sequence 2 due to known acquisition issues\footnote{http://www.cas.kth.se/COLD/bugs.php} and Saarbr\"ucken Sunny, for which we take part B since sunny sequences for part A are not available. As in the previous section, we consider the same 4 classes shared between the sequences, reporting the average accuracy per class. In these experiments we consider both AlexNet and ResNet comparing our approach with baseline models obtained adding traditional BN layers to the same architectures.

Firstly, we consider different lighting conditions, \ie we assume that the domain shift is due to changes of illuminations. To this extent we train the network on sequences of the same laboratory, training on two lighting conditions (\eg \textit{sunny} and \textit{cloudy}) and testing on the third (\eg \textit{night}). The results are reported in Table \ref{tab:cold-dg-lab}. 
\begin{figure}[t]
  \centering
  \includegraphics[width=0.4\textwidth,trim={0cm 0.6cm 0cm 9.5cm},clip]{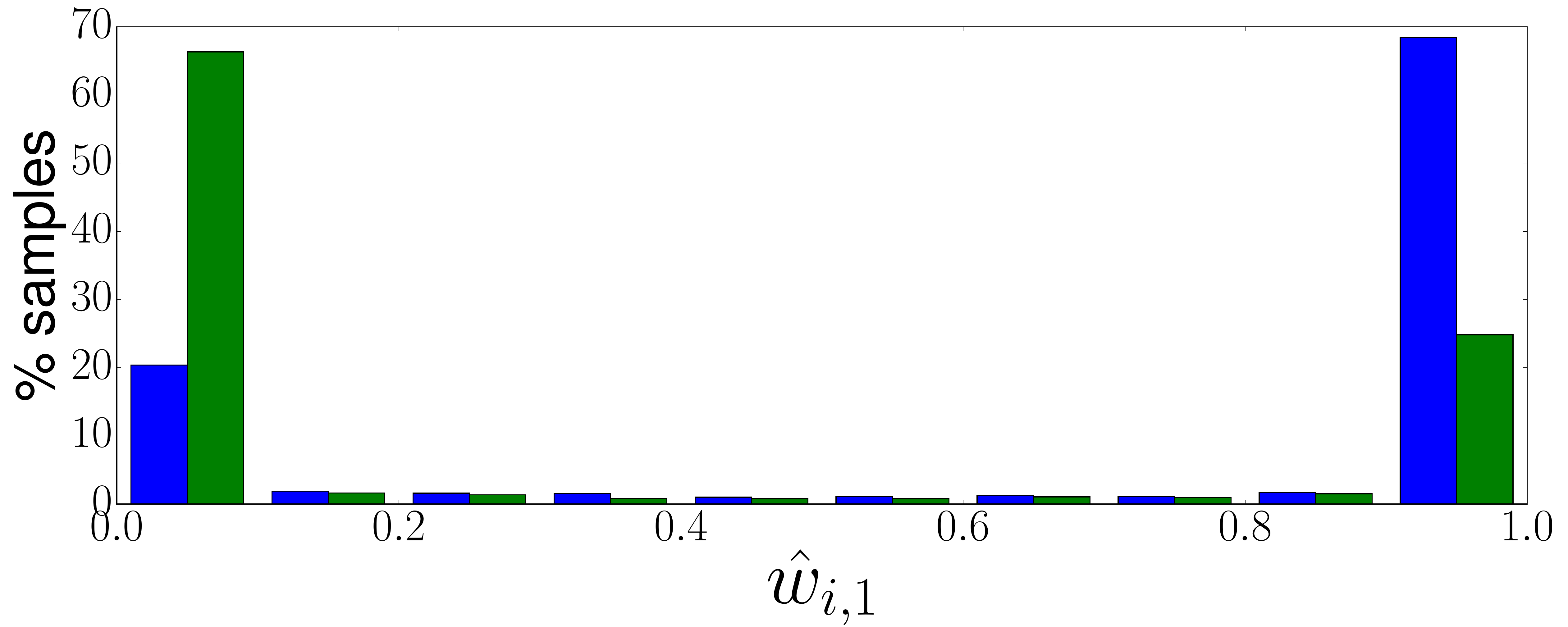}
\vspace{-6pt}
  \caption{Distribution of the values of the weights computed with AlexNet+WBN for the scenario Lj.N as target in Table \ref{tab:cold-dg-lig}. Different colors represent different original source domains. \vspace{-15pt}}
   \label{fig:vis-cold}
\end{figure}
{As expected, when knowledge about domains is available (WBN$ ^*$), improved classification accuracy can be obtained, in general, with respect to a domain agnostic classifier. Interestingly, for both networks the result of our approach without domain priors is either comparable or surpasses the baseline in almost all settings. This suggests that the network is able to latently discover clusters of samples and effectively using this information for learning robust classification models. 
}

Secondly, we perform a similar analysis to Table \ref{tab:cold-dg-lab} but considering changes of robotic platform\slash{}environment. We keep constant the lighting condition, training on two laboratories and testing on the third. Table \ref{tab:cold-dg-lig} shows the obtained results. Again, in most cases exploiting domain priors brings benefits in term of performances, for both networks. The results of Tables \ref{tab:cold-dg-lab} and \ref{tab:cold-dg-lig} show that the benefits of our WBN layer, with and without domain loss, are not limited to a particular type of domain shift (\ie changes in robots, environment or illumination condition), demonstrating that our approach provides a general and effective strategy to address domain variations. 
{In both experiments, there are few cases in which standard BN achieves comparable or slightly superior results w.r.t. WBN. A possible reason is that in some situations the ability of our model to generalize to novel settings may be hindered by the small number or by the specific characteristics of the available source domains.

{In order to verify the ability of WBN to discover latent domains, Fig. \ref{fig:vis-cold} shows the distribution of the values $\hat{w}_{i,j}$ computed for the images of the original source domains associated to one of the experiments in Table \ref{tab:cold-dg-lig}. The plots associated to other experiments are similar and we do not report them due to lack of space. Since we consider two latent domains in these experiments and $\hat{w}_{i,1}+\hat{w}_{i,2}=1$, we report only the values computed for $\hat{w}_{i,1}$
. Different colors represent the original source domains. As the figure shows, the lateral branch computes different assignments for the samples of the different original source domains. As a result, the latent source domains extracted by WBN tend to correspond to the original source domains used by WBN$^*$.} 



\begin{table}[t]
			\caption{DG accuracy on COLD over different lighting conditions. First row indicates the target sequence, with the first letters denoting the laboratory and the last the illumination condition (C=Cloudy, S=Sunny, N=Night). Vertical lines separate domains of the same laboratory.\vspace{-5pt}} 
		\centering
		\scalebox{.88}{
		\begin{tabular}{| c | l@{\hskip3pt} | c@{\hskip5pt}  c@{\hskip5pt}  c@{\hskip5pt} | c@{\hskip5pt} c@{\hskip5pt} c@{\hskip5pt} | c@{\hskip5pt} c@{\hskip5pt} c@{\hskip5pt} | c | } 
			\hline
			Net& Norm. & Fr.C & Fr.N & Fr.S & Lj.C & Lj.N & Lj.S &Sa.C &Sa.N & Sa.S& avg.\\
			\hline
					
 \multirow{3}{*}{\rotatebox[origin=c]{90}{AlexNet}}&BN&97.3	&89.1	&97.4	&92.9	&64.4	&94.2	&75.6	&69.7	&44.0	&80.5\\
 &WBN&\textbf{98.1}	&91.3	&97.1	&93.1	&65.1	&94.1	&\textbf{77.7}	&68.8	&\textbf{50.2}	&81.7\\
 &WBN$ ^*$&97.1	&\textbf{91.9}	&\textbf{98.0}	&\textbf{93.9}	&\textbf{65.6}	&\textbf{95.0}	&77.2	&\textbf{69.9}	&49.9	&\textbf{82.1}\\
\hline
\multirow{3}{*}{\rotatebox[origin=c]{90}{ResNet}}&BN&97.7	&\textbf{82.2}	&90.7	&89.5	&61.2	&90.3	&70.7	&73.0	&\textbf{38.7}	&77.1\\
&WBN&\textbf{98.1}	&81.8	&\textbf{94.1}	&94.5	&61.7	&93.7	&75.8	&76.9	&37.8	&79.4\\
&WBN$ ^*$&97.9	&81.3	&93.4	&\textbf{94.7}	&\textbf{65.1}	&\textbf{94.6}	&\textbf{78.1}	&\textbf{76.5}	&38.5	&\textbf{80.0}\\
            \hline
		\end{tabular}
        }
		\label{tab:cold-dg-lab}
        \vspace{-8pt}
\end{table}

\begin{table}[t]
			\caption{DG accuracy on COLD over different environments\slash{}sensors. First row indicates the target sequence, with the first letters denoting the laboratory and the last the illumination condition (C=Cloudy, S=Sunny, N=Night). Vertical lines separate domains with same illumination condition.\vspace{-5pt}} 
		\centering
		\scalebox{.88}{
		\begin{tabular}{| c | l@{\hskip3pt} | c@{\hskip5pt}  c@{\hskip5pt}  c@{\hskip5pt} | c@{\hskip5pt} c@{\hskip5pt} c@{\hskip5pt} | c@{\hskip5pt} c@{\hskip5pt} c@{\hskip5pt} | c | } 
			\hline
			Net& Norm.& Fr.C & Sa.C & Lj.C & Fr.N & Sa.N & Lj.N &Fr.S &Lj.S & Sa.S& avg.\\
			\hline
					
	\multirow{3}{*}{\rotatebox[origin=c]{90}{AlexNet}}&BN&\textbf{26.0}	&38.4	&\textbf{34.4}	&27.9	&26.6	&33.1	&28.8	&34.2	&25.1	&30.5\\
&WBN&25.8	&38.2	&33.0	&\textbf{29.4}	&26.6	&34.8	&30.3	&36.9	&25.1	&31.1\\
&WBN$ ^*$&25.9	&\textbf{40.3}	&33.4	&28.0	&\textbf{27.6}	&\textbf{34.9}	&\textbf{31.5}	&\textbf{44.3}	&\textbf{28.6}	&\textbf{32.7}\\
\hline
\multirow{3}{*}{\rotatebox[origin=c]{90}{ResNet}}&BN&\textbf{37.9}	&\textbf{40.9}	&39.3	&30.8	&48.3	&41.2	&30.6	&\textbf{40.6}	&27.6	&37.5\\
&WBN&37.3	&39.5	&\textbf{42.6}	&40.4	&51.8	&41.0	&33.8	&39.6	&\textbf{30.8}	&39.6\\
&WBN$ ^*$&36.6	&40.3	&40.0	&\textbf{41.2}	&\textbf{56.2}	&\textbf{45.2}	&\textbf{35.4}	&39.4	&25.6	&\textbf{40.0}\\
            \hline
		\end{tabular}
        }
		\label{tab:cold-dg-lig}
        \vspace{-15pt}
\end{table}

{In another series of experiments we consider the scenario where both illumination and laboratory change. We performed 27 different experiments, corresponding to the case where Saarbr\"ucken is considered as target domain. Figure \ref{fig:hist-81-cold} report the histogram of the gains in accuracy of our approach AlexNet+WBN* w.r.t. AlexNet+BN. As shown in Fig. \ref{fig:hist-81-cold}, in most of the cases our model leads to an increase in accuracy between 1-5\%. In only 5 out of 27 experiments, our model does not produce benefits.} 

\begin{figure}[t]
\centering
\includegraphics[width=0.8\columnwidth,trim={0cm 0.48cm 0cm 8.8cm},clip]{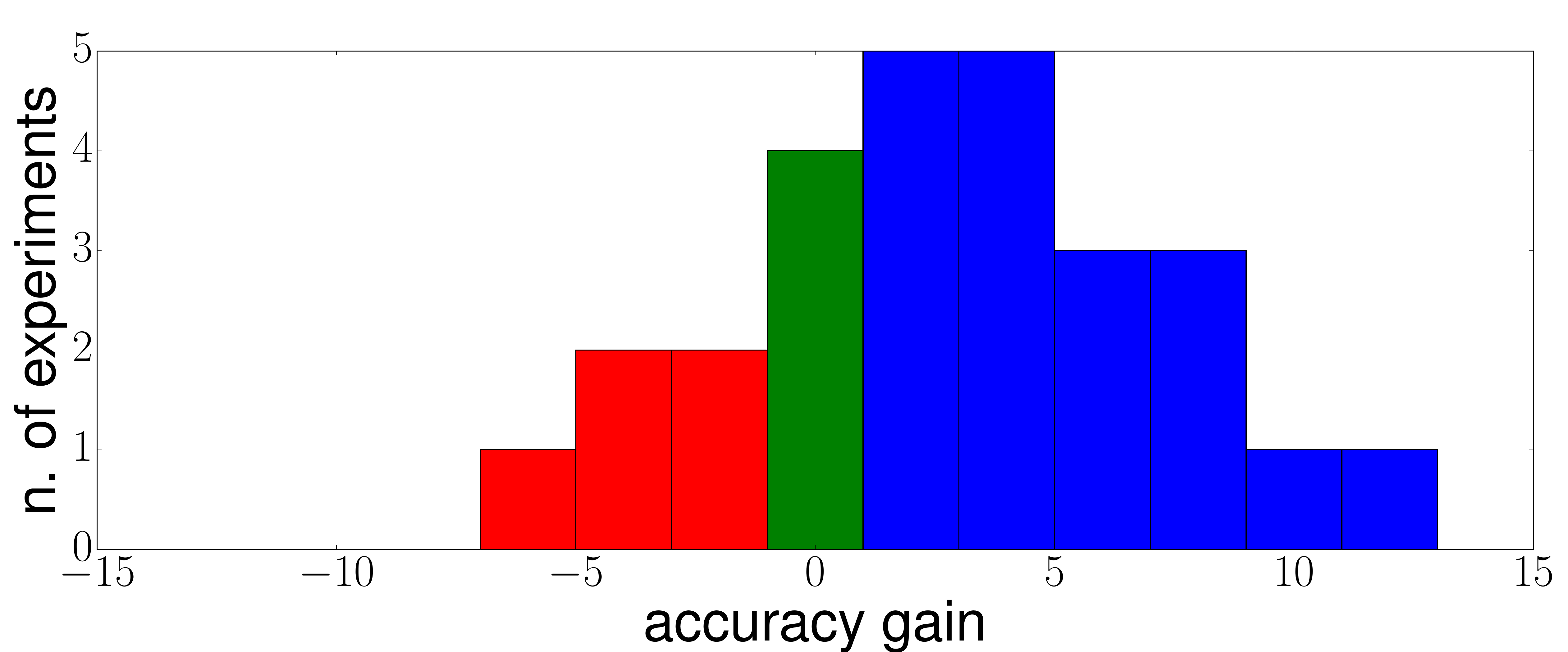}
\vspace{-6pt}
\caption{COLD dataset: results at varying both laboratory and illumination  conditions considering Saarbr\"ucken as target. Histogram of the accuracy gains of AlexNet+WBN* w.r.t. AlexNet+BN. Colors indicate positive (blue) and negative (red) gains and comparable performance (green).\vspace{-20pt}}
   \label{fig:hist-81-cold}
\end{figure}

In order to compare our model with the state-of-the art approaches, we  consider the VPC dataset. VPC has been used in previous works to test the DG abilities of different methods. Following the standard experimental protocol of \cite{wu2009visual}, we evaluate our model using 5 houses for training and 1 for test, averaging the results between the 6 configurations. For each house we report the average accuracy per class. 
Table \ref{tab:vpc-comparison} compares the result of our models with baseline deep architectures, with and without traditional BN layers. We consider both the case where domain information is available (WBN$ ^*$) and where it is not (WBN). Analogously to what observed in the experiments on COLD dataset, the accuracy increases when WBN is adopted, both in case of AlexNet and ResNet architectures. 
Interestingly, having domain priors during training produce a boost of performances for ResNet, while for AlexNet this is not the case. This suggests that different features have a different impact on our model. Features of the very last layers, as in AlexNet, may not be enough domain discriminative, especially in case of limited shift within the source domains. In those cases, a soft-assignment can provide a more effective strategy for clustering samples}.

\begin{table}[t]
			\caption{VPC dataset: average accuracy per class. \vspace{-5pt}} 
		\centering
		\scalebox{.9}{
		\begin{tabular}{| l | c c c c c c | c |} 
			\hline
			Net & H1 & H2& H3&H4 &H5 &H6& avg.\\
            \hline
            
			AlexNet & 49.8 & 53.4&	49.2&	64.4 &41.0	&43.4	& 50.2\\
AlexNet + BN& 54.5 &  \textbf{54.6} &  55.6 &  69.7   &  41.8 &  45.9 & 53.7\\
AlexNet + WBN &\textbf{54.7}& 51.9&	\textbf{61.8}&	\textbf{70.6}&	43.9& 46.5 &\textbf{54.9}\\
AlexNet + WBN$ ^*$ &53.5& \textbf{54.6} & 55.7 &	68.1&	\textbf{44.3}& \textbf{49.9} & 54.3 \\
\hline
ResNet & 55.8&  47.4 &  64.0 &  69.9 &  42.8 & 50.4 & 55.0\\
ResNet + WBN &55.7 &49.5 &\textbf{64.7 }&\textbf{70.2} &42.1 &\textbf{52.0} &55.7\\
ResNet + WBN$ ^*$& \textbf{56.8} & \textbf{50.9} & 64.1 &  69.3  &   \textbf{45.1} & 51.6&\textbf{56.5}\\
            \hline
            
		\end{tabular}
        }
		\label{tab:vpc-comparison}
         \vspace{-7pt}
\end{table}

Finally, Table \ref{tab:vpc-sota} compares the results obtained with our approach with those of state-of-the-art methods. Specifically we consider the method in \cite{wu2009visual}, where SIFT \cite{lowe2004distinctive} 
and CENTRIST (CE) features \cite{wu2011centrist} are provided as input to a nearest neighbor classifier, and the approach in \cite{fazl2012histogram}, where the same classifier is employed but using Histogram of Oriented Uniform Patterns (HOUP) as input. For sake of completeness, we also report the results obtained by exploiting also the temporal information between images. For this setting, we report the performances of the CENTRIST-based approach of \cite{wu2011centrist} coupled with Bayesian Filtering (BF) and the results of \cite{yang2012object} which used again a Bayesian Filter together with object templates. 
As shown in the Table, applying deep-learning techniques already guarantees an increase in performances of about $4\%$ with respect to the state-of-the-art. Introducing WBN inside the network, allows a further accuracy gain. 

\begin{table}[t]
			\caption{VPC dataset: comparison with state of the art.\vspace{-5pt}} 
		\centering
		\scalebox{.95}{
		\begin{tabular}{| l | @{\hskip2pt}c@{\hskip3pt}  @{\hskip2pt}c@{\hskip3pt}  @{\hskip2pt}c@{\hskip3pt} | @{\hskip3pt}c@{\hskip3pt} | @{\hskip3pt}c@{\hskip3pt} | @{\hskip3pt}c@{\hskip2pt} @{\hskip2pt}c@{\hskip2pt} @{\hskip2pt}c@{\hskip3pt} | @{\hskip3pt}c@{\hskip2pt}  @{\hskip2pt}c@{\hskip3pt}|} 
			\hline
			Method & &\cite{wu2009visual}& & \cite{fazl2012histogram} & \cite{yang2012object} & \multicolumn{3}{@{\hskip3pt}c|@{\hskip3pt}}{AlexNet}&\multicolumn{2}{@{\hskip3pt}c@{\hskip3pt}|}{ResNet}\\ \hline
            Config. & SIFT & CE & BF & - & - & Base & BN &WBN$^*$& BN &WBN$^*$\\
            \hline
            Acc.& 35.0&41.9&45.6&45.9&50.0&50.2&53.7&54.3&55.0&\textbf{56.5}\\
            \hline
		\end{tabular}
        }
		\label{tab:vpc-sota}
        \vspace{-15pt}
\end{table}

\subsubsection{Large scale experiments}
\label{exp:large-scale}
{In this section we show the results obtained when our method is applied to a large scale dataset of outdoor scenes, \ie the SPED dataset. In order to employ SPED as a DG benchmark, we split the dataset in two sets, February and August, considering the months of data acquisition. Since no other automatic training data splits are possible using timestamps, in these experiments we do not use domain supervision and only consider WBN with two latent domains. The choice of having two domains is motivated by the fact that the dataset contains images collected at different times of the day and thus we assume that the latent domains automatically discovered by our method correspond to "night" and "day".} 

\begin{table}[t]
			\caption{SPED dataset: comparison of different models.\vspace{-5pt}} 
		\centering
		\scalebox{1.0}{
		\begin{tabular}{| l | c  c  c | c  c  c |} 
			\hline
			Net & \multicolumn{3}{c|}{AMOSNet}  & \multicolumn{3}{c|}{AlexNet}\\
            \hline
            Config. & Base & BN & WBN & Base & BN & WBN\\
            \hline
            February-to-August & 83.7 & 88.8 & \textbf{90.3} &83.6&88.9 & \textbf{90.5 }\\
            August-to-February & 71.2&82.7&\textbf{86.1}&73.9&83.1&\textbf{87.0}\\
            \hline
		\end{tabular}
        }
		\label{tab:sped}
        \vspace{-20pt}
\end{table}

{Results are shown in Table \ref{tab:sped}. WBN provides a clear gain in all considered settings and for all considered architectures. The improvement of 4\% obtained in the case "August-to-February" for both networks is remarkable given the very large number of classes and the lack of domain supervision.}

\section{DISCUSSION AND CONCLUSIONS}
We presented a deep learning model for addressing DG in the context of semantic place categorization. Our approach exploits a weighted formulation of BN to learn robust classifiers which can be applied to previously unseen target domains.
Our experiments demonstrated that, by adopting our WBN layers, place categorization accuracy can be significantly increased and 
state-of-the-art performance on the VPC benchmark can be obtained. The effectiveness of our method is also confirmed by experiments on a large scale dataset of outdoor scenes.
In future works we plan to further explore our DG framework, analyzing the impact of computing different weighting vectors for different layers
and exploiting different structures for the lateral network branch.  
{We believe that the generality of our framework also permits to exploit additional sources of knowledge, such as web images \cite{song2015robot} and synthetic data \cite{mccormac2017scenenet}. For instance, we can think of building a large training set of images through web queries and employ the unsupervised version of our algorithm to infer the latent source domains useful for DG.\vspace{-0pt}} 
\bibliographystyle{IEEEtran}
\bibliography{IEEEabrv,root}

\begin{thebibliography}{10}
\providecommand{\url}[1]{#1}
\csname url@rmstyle\endcsname
\providecommand{\newblock}{\relax}
\providecommand{\bibinfo}[2]{#2}
\providecommand\BIBentrySTDinterwordspacing{\spaceskip=0pt\relax}
\providecommand\BIBentryALTinterwordstretchfactor{4}
\providecommand\BIBentryALTinterwordspacing{\spaceskip=\fontdimen2\font plus
\BIBentryALTinterwordstretchfactor\fontdimen3\font minus
  \fontdimen4\font\relax}
\providecommand\BIBforeignlanguage[2]{{%
\expandafter\ifx\csname l@#1\endcsname\relax
\typeout{** WARNING: IEEEtran.bst: No hyphenation pattern has been}%
\typeout{** loaded for the language `#1'. Using the pattern for}%
\typeout{** the default language instead.}%
\else
\language=\csname l@#1\endcsname
\fi
#2}}

\bibitem{he2016deep}
K.~He, X.~Zhang, S.~Ren, and J.~Sun, ``Deep residual learning for image
  recognition,'' in \emph{CVPR}, 2016.

\bibitem{xu2017multi}
D.~Xu, E.~Ricci, W.~Ouyang, X.~Wang, and N.~Sebe, ``Multi-scale continuous crfs
  as sequential deep networks for monocular depth estimation,'' \emph{CVPR},
  2017.

\bibitem{porzi2017learning}
L.~Porzi, S.~R. Bul{\'o}, A.~Penate-Sanchez, E.~Ricci, and F.~Moreno-Noguer,
  ``Learning depth-aware deep representations for robotic perception,''
  \emph{IEEE RA-L}, vol.~2, no.~2, pp. 468--475, 2017.

\bibitem{wu2009visual}
J.~Wu, H.~I. Christensen, and J.~M. Rehg, ``Visual place categorization:
  Problem, dataset, and algorithm,'' in \emph{IROS}, 2009.

\bibitem{stachniss2006speeding}
C.~Stachniss, O.~M. Mozos, and W.~Burgard, ``Speeding-up multi-robot
  exploration by considering semantic place information,'' in \emph{ICRA},
  2006.

\bibitem{kostavelis2015semantic}
I.~Kostavelis and A.~Gasteratos, ``Semantic mapping for mobile robotics tasks:
  A survey,'' \emph{Robotics and Autonomous Systems}, vol.~66, pp. 86--103,
  2015.

\bibitem{wu2011centrist}
J.~Wu and J.~M. Rehg, ``Centrist: A visual descriptor for scene
  categorization,'' \emph{IEEE T-PAMI}, vol.~33, no.~8, pp. 1489--1501, 2011.

\bibitem{fazl2012histogram}
E.~Fazl-Ersi and J.~K. Tsotsos, ``Histogram of oriented uniform patterns for
  robust place recognition and categorization,'' \emph{IJRR}, vol.~31, no.~4,
  pp. 468--483, 2012.

\bibitem{urvsivc2016part}
P.~Ur{\v{s}}i{\v{c}}, A.~Leonardis, M.~Kristan, \emph{et~al.}, ``Part-based
  room categorization for household service robots,'' in \emph{ICRA}, 2016.

\bibitem{mancini2017learning}
M.~Mancini, S.~Rota~Bul{\`o}, E.~Ricci, and B.~Caputo, ``Learning deep nbnn
  representations for robust place categorization,'' \emph{IEEE RA-L}, vol.~2,
  no.~3, pp. 1794--1801, 2017.

\bibitem{pronobis2010realistic}
A.~Pronobis, B.~Caputo, P.~Jensfelt, and H.~I. Christensen, ``A realistic
  benchmark for visual indoor place recognition,'' \emph{Robotics and
  autonomous systems}, vol.~58, no.~1, pp. 81--96, 2010.

\bibitem{prasath2012transfer}
S.~Prasath~Elango, T.~Tommasi, and B.~Caputo, ``Transfer learning of visual
  concepts across robots: A discriminative approach,'' Idiap, Tech. Rep., 2012.

\bibitem{costante2013transfer}
G.~Costante, T.~A. Ciarfuglia, P.~Valigi, and E.~Ricci, ``A transfer learning
  approach for multi-cue semantic place recognition,'' in \emph{IROS}, 2013.

\bibitem{kira2014transfer}
Z.~Kira, ``Transfer of sparse coding representations and object classifiers
  across heterogeneous robots,'' in \emph{IROS}, 2014.

\bibitem{muandet2013domain}
K.~Muandet, D.~Balduzzi, and B.~Sch{\"o}lkopf, ``Domain generalization via
  invariant feature representation,'' in \emph{ICML}, 2013.

\bibitem{khosla2012undoing}
A.~Khosla, T.~Zhou, T.~Malisiewicz, A.~A. Efros, and A.~Torralba, ``Undoing the
  damage of dataset bias,'' in \emph{ECCV}, 2012.

\bibitem{carlucci2017just}
F.~M. Carlucci, L.~Porzi, B.~Caputo, E.~Ricci, and S.~R. Bul{\`o}, ``Just dial:
  Domain alignment layers for unsupervised domain adaptation,'' in
  \emph{ICIAP}, 2017.

\bibitem{li1603revisiting}
Y.~Li, N.~Wang, J.~Shi, J.~Liu, and X.~Hou, ``Revisiting batch normalization
  for practical domain adaptation,'' 2017.

\bibitem{pronobis2009ijrr}
A.~Pronobis and B.~Caputo, ``{COLD}: {CO}sy {L}ocalization {D}atabase,''
  \emph{IJRR}, vol.~28, no.~5, pp. 588--594, 2009.

\bibitem{chen2017deep}
Z.~Chen, A.~Jacobson, N.~Sunderhauf, B.~Upcroft, L.~Liu, C.~Shen, I.~Reid, and
  M.~Milford, ``Deep learning features at scale for visual place recognition,''
  \emph{arXiv preprint arXiv:1701.05105}, 2017.

\bibitem{kanji2015cross}
T.~Kanji, ``Cross-season place recognition using nbnn scene descriptor,'' in
  \emph{IROS}, 2015.

\bibitem{viswanathan2016vision}
A.~Viswanathan, B.~R. Pires, and D.~Huber, ``Vision-based robot localization
  across seasons and in remote locations,'' in \emph{ICRA}, 2016.

\bibitem{carlevaris2014learning}
N.~Carlevaris-Bianco and R.~M. Eustice, ``Learning visual feature descriptors
  for dynamic lighting conditions,'' in \emph{IROS}, 2014.

\bibitem{lu2015robustness}
Y.~Lu and D.~Song, ``Robustness to lighting variations: An rgb-d indoor visual
  odometry using line segments,'' in \emph{IROS}, 2015.

\bibitem{mount20162d}
J.~Mount and M.~Milford, ``2d visual place recognition for domestic service
  robots at night,'' in \emph{ICRA}, 2016.

\bibitem{upcroft2014lighting}
B.~Upcroft, C.~McManus, W.~Churchill, W.~Maddern, and P.~Newman, ``Lighting
  invariant urban street classification,'' in \emph{ICRA}, 2014.

\bibitem{milford2012seqslam}
M.~J. Milford and G.~F. Wyeth, ``Seqslam: Visual route-based navigation for
  sunny summer days and stormy winter nights,'' in \emph{ICRA}, 2012.

\bibitem{tsukamoto2015self}
T.~Tsukamoto and K.~Tanaka, ``Self-localization using visual experience across
  domains,'' \emph{arXiv preprint arXiv:1509.07618}, 2015.

\bibitem{pronobis2006discriminative}
A.~Pronobis, B.~Caputo, P.~Jensfelt, and H.~I. Christensen, ``A discriminative
  approach to robust visual place recognition,'' in \emph{IROS}, 2006.

\bibitem{zhou2014learning}
B.~Zhou, A.~Lapedriza, J.~Xiao, A.~Torralba, and A.~Oliva, ``Learning deep
  features for scene recognition using places database,'' in \emph{NIPS}, 2014.

\bibitem{liao2016understand}
Y.~Liao, S.~Kodagoda, Y.~Wang, L.~Shi, and Y.~Liu, ``Understand scene
  categories by objects: A semantic regularized scene classifier using
  convolutional neural networks,'' in \emph{ICRA}, 2016.

\bibitem{yang2012object}
H.~Yang and J.~Wu, ``Object templates for visual place categorization.'' in
  \emph{ACCV}, 2012, pp. 470--483.

\bibitem{luo2007svm}
J.~Luo, A.~Pronobis, and B.~Caputo, ``Svm-based transfer of visual knowledge
  across robotic platforms,'' 2007.

\bibitem{xu2014exploiting}
Z.~Xu, W.~Li, L.~Niu, and D.~Xu, ``Exploiting low-rank structure from latent
  domains for domain generalization,'' in \emph{ECCV}, 2014.

\bibitem{ioffe2015batch}
S.~Ioffe and C.~Szegedy, ``Batch normalization: Accelerating deep network
  training by reducing internal covariate shift,'' in \emph{ICML}, 2015.

\bibitem{DBLP:journals/corr/WulfmeierBP17}
M.~Wulfmeier, A.~Bewley, and I.~Posner, ``Addressing appearance change in
  outdoor robotics with adversarial domain adaptation,'' \emph{arXiv preprint
  arXiv:1703.01461}.

\bibitem{carlucci2017autodial}
F.~M. Carlucci, L.~Porzi, B.~Caputo, E.~Ricci, and S.~R. Bul{\`o}, ``Autodial:
  Automatic domain alignment layers,'' 2017.

\bibitem{jacobs2007consistent}
N.~Jacobs, N.~Roman, and R.~Pless, ``Consistent temporal variations in many
  outdoor scenes,'' in \emph{CVPR}, 2007.

\bibitem{krizhevsky2012imagenet}
A.~Krizhevsky, I.~Sutskever, and G.~E. Hinton, ``Imagenet classification with
  deep convolutional neural networks,'' in \emph{NIPS}, 2012.

\bibitem{deng2009imagenet}
J.~Deng, W.~Dong, R.~Socher, L.-J. Li, K.~Li, and L.~Fei-Fei, ``Imagenet: A
  large-scale hierarchical image database,'' in \emph{CVPR}, 2009.

\bibitem{jia2014caffe}
Y.~Jia, E.~Shelhamer, J.~Donahue, S.~Karayev, J.~Long, R.~Girshick,
  S.~Guadarrama, and T.~Darrell, ``Caffe: Convolutional architecture for fast
  feature embedding,'' in \emph{ACM Multimedia}, 2014.

\bibitem{simon2016cnnmodels}
M.~Simon, E.~Rodner, and J.~Denzler, ``Imagenet pre-trained models with batch
  normalization,'' \emph{arXiv preprint arXiv:1612.01452}, 2016.

\bibitem{lowe2004distinctive}
D.~G. Lowe, ``Distinctive image features from scale-invariant keypoints,''
  \emph{IJCV}, vol.~60, no.~2, pp. 91--110, 2004.

\bibitem{song2015robot}
S.~Song, L.~Zhang, and J.~Xiao, ``Robot in a room: Toward perfect object
  recognition in closed environments,'' \emph{CoRR, abs/1507.02703}, 2015.

\bibitem{mccormac2017scenenet}
J.~McCormac, A.~Handa, S.~Leutenegger, and A.~J. Davison, ``Scenenet rgb-d: Can
  5m synthetic images beat generic imagenet pre-training on indoor
  segmentation,'' in \emph{CVPR}, 2017.

\end{thebibliography}

\end{document}